\documentclass[letterpaper]{article}
\usepackage{aaai2027}
\usepackage[hyphens]{url}
\usepackage{graphicx}
\usepackage{natbib}
\usepackage{caption}
\usepackage{amsmath,amssymb,amsfonts,mathtools,bm}
\usepackage{nicefrac}
\usepackage{booktabs}
\usepackage{multirow}
\usepackage{xcolor}
\usepackage{colortbl}
\usepackage{algorithm}
\usepackage{algpseudocode}
\usepackage{cleveref}


\ifdefined{/TemplateVersion (2027.1)}\fi  
\nocopyright

\title{Semantic Reasoning Denoising:\\
       Correcting Language Model Reasoning with Semantic Operators}
\author{Yujiao Yang}
\affiliations{
Dalian University of Technology\\
yjyang@mail.dlut.edu.cn
}

\begin{document}
\maketitle

\begin{abstract}
Large language models can produce fluent reasoning traces whose local semantic
errors propagate to an incorrect conclusion, while unconstrained
self-correction may preserve, amplify, or introduce errors. Existing diffusion
language models provide iterative refinement, but usually define noise as token
masking or replacement rather than as errors in the reasoning process. We
present Semantic Reasoning Denoising (SRD), an operatorized Markov denoising
method for natural-language reasoning trajectories. SRD represents
semantic noise with executable error operators that describe the error type,
its location, and the corrupted and repaired propositions. Composing these
operators constructs progressively noisier states. During training, the model
learns to identify the semantic noise active in the current trajectory and to
reconstruct the paired adjacent lower-noise state. During inference,
noise-level-aware denoising repeatedly predicts an inverse operator and checks
whether it is applicable, so each executed update makes a localized move toward a stable
trajectory. Across six in-domain benchmarks spanning mathematics, code,
knowledge, and commonsense, SRD improves the strongest same-backbone baseline
by 3.2 points on average. On seven cross-dataset transfer targets, it remains
competitive with Llama-3-8B-Instruct and improves the strongest Qwen3-8B
baseline average by 2.9 points. Analyses of noise sources, objectives, and
denoising depth further show that structured semantic-noise prediction and
iterative operator execution are central to the improvement.
\end{abstract}


\section{Introduction}
\label{sec:intro}

\begin{figure*}[t]
\centering
\includegraphics[width=0.97\textwidth]{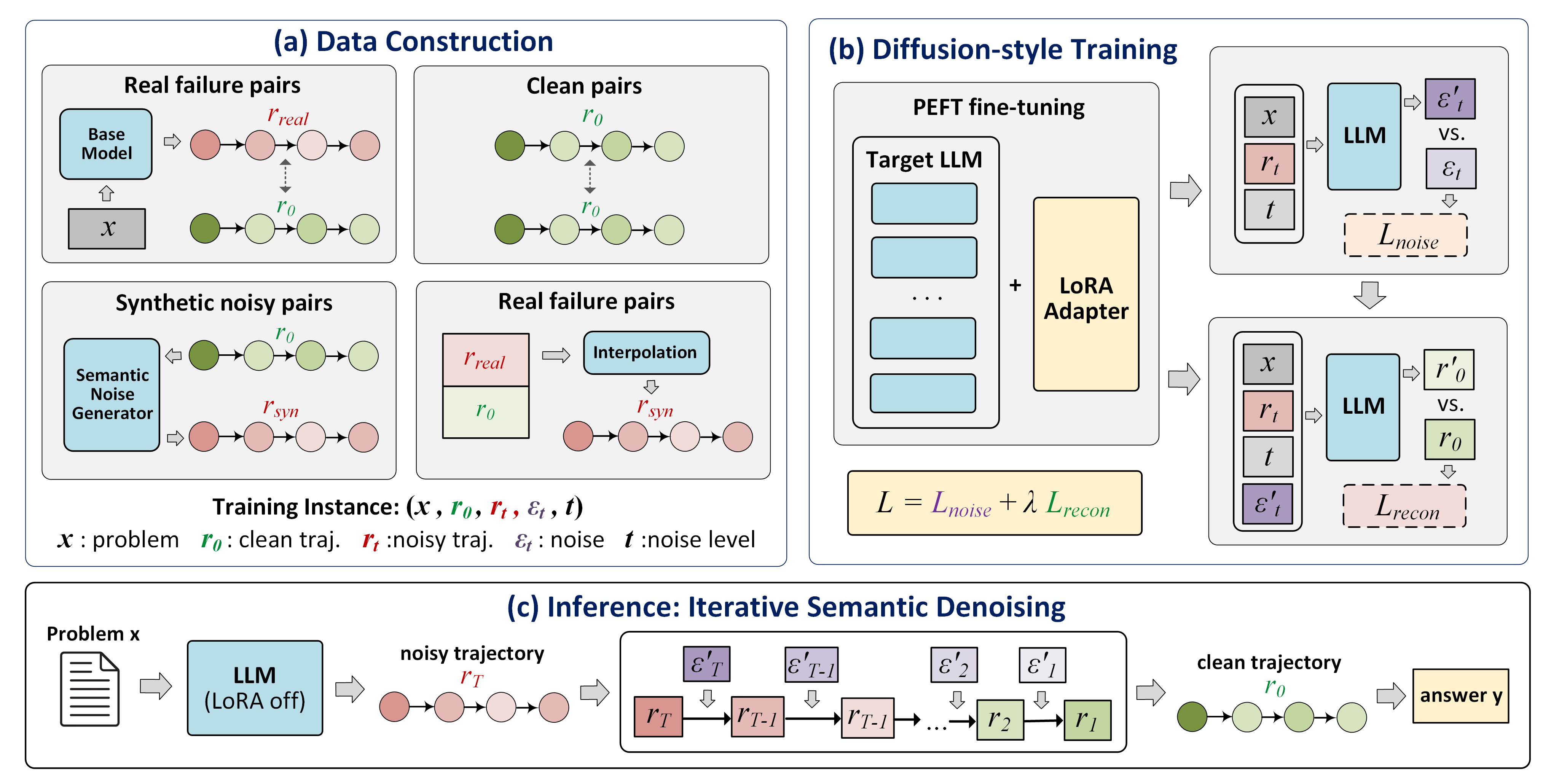}
\caption{Overview of SRD. \textbf{(a)} Real, synthetic, and interpolated
failures are aligned into executable semantic-noise operators and adjacent
states. \textbf{(b)} A LoRA denoiser predicts the active operator and adjacent
lower-noise state. \textbf{(c)} From the base CoT, applicability-checked inverse
operators iteratively recover the final answer.}
\label{fig:overview}
\end{figure*}

Large language models (LLMs) have made remarkable progress on complex reasoning
tasks, yet most approaches still focus on direct generation and lack an explicit
mechanism for progressively correcting intermediate reasoning states with local
errors. Diffusion models provide an instructive process-level perspective:
generation proceeds through progressive corruption and reverse recovery, a
structure naturally suited to describing iterative refinement from erroneous to
correct reasoning~\citep{ho2020denoising,sohl2015deep}.

Existing diffusion language models typically define noise through masking,
replacement, or other token-level perturbations, and generate text by
progressively recovering the original sequence~\citep{austin2021structured,nie2025large}.
Such perturbations suit general text generation but do not directly capture
reasoning errors, which often involve incorrect intermediate conclusions, misuse
of conditions, or broken logical dependencies. Token-level diffusion therefore
primarily recovers textual form rather than the underlying reasoning process.

To this end, we propose \textbf{SRD} (Semantic Reasoning Denoising), an
operatorized Markov denoising method for natural-language reasoning
trajectories. SRD borrows the progressive-corruption and reverse-recovery
mechanisms of diffusion models, represents semantic noise with executable error
operators, and composes those operators to construct a sequence of coherent
corrupted reasoning states. Conditioned on a normalized process coordinate, the model
learns ordered local inverse transitions that progressively recover the
reasoning trajectory.

SRD comprises three stages (\cref{fig:overview}): training-sample construction,
diffusion-style training, and iterative inference. In the data-construction
stage, we jointly use real error trajectories, synthetic noised trajectories,
and interpolated trajectories between erroneous and correct ones to cover
different noise levels. During training, the model learns both to identify the
semantic noise in the current trajectory---represented by an executable error
operator that specifies the error type, location, and corrupted and repaired
content---and, conditioned on this diagnosis, to recover a lower-noise or clean
reasoning state. During inference, several rounds of noise-level-aware semantic
denoising progressively refine the trajectory. Each round predicts one inverse
operator and checks its applicability, yielding an explicit adjacent transition instead of
unconstrained regeneration.

Our experiments on $13$ tasks spanning mathematical reasoning, code generation,
and general understanding show that a few thousand transition pairs suffice to
train the SRD adapter. Across six in-domain benchmarks, SRD averages 67.5,
outperforming the strongest same-backbone baseline by 3.2 points and leading it
on all six tasks; SRD is also the best same-backbone method on five. It remains
competitive on Llama transfer and gives the best Qwen3 transfer average. These
results support the effectiveness of an operatorized diffusion-style process
over reasoning trajectories.

\paragraph{Contributions.}
In summary, our contributions are as follows:
\begin{itemize}
  \item We formulate reasoning refinement as an operatorized semantic denoising
    process, in which executable and compositional error operators define
    explicit transitions between coherent reasoning states.
  \item We propose SRD, which constructs trajectories at different semantic-noise
    levels and learns ordered local inverse transitions for iterative reasoning
    recovery.
  \item Experiments across math, code, and general reasoning demonstrate strong
    in-domain performance, sample efficiency, and cross-dataset transfer.
\end{itemize}


\section{Related Work}
\label{sec:related}

\paragraph{Reasoning generation, refinement, and correction.}
Chain-of-thought prompting~\citep{wei2022chain} elicits intermediate reasoning
steps, while STaR~\citep{zelikman2022star} bootstraps training from
model-generated rationales. Self-consistency~\citep{wang2023selfconsistency}
subsequently aggregates multiple sampled chains, and
Self-Refine~\citep{madaan2023selfrefine} iteratively revises an answer using
self-feedback. More recent methods learn correction behavior through recursive
self-improvement in RISE~\citep{qu2024rise}, reinforcement learning in
SCoRe~\citep{kumar2024score}, supervised and reinforcement-learned
self-verification in S$^2$R~\citep{ma2025s2r}, or executable pseudo-program
feedback in ProgCo~\citep{song2025progco}. Recent analysis also distinguishes
preserving correct answers from repairing incorrect ones
\citep{yang2025confidence}. Accordingly, our experiments compare standard SFT
and STaR as generation-oriented baselines, and Self-Refine, RISE, and SCoRe as
strong correction baselines. Unlike free-form response revision, SRD identifies
typed semantic noise and applies an executable local transition, preserving
unaffected portions of the trajectory.

\paragraph{Diffusion language models and diffusion-style reasoning.}
Discrete diffusion originally models text corruption through token masking or
replacement~\citep{austin2021structured}. Diffusion of
Thoughts~\citep{ye2024diffusion} applies token-level denoising to
chain-of-thought generation. LLaDA-8B~\citep{nie2025large} and
Dream-7B~\citep{ye2025dream} subsequently scale discrete diffusion to
general-purpose language models; we report them as diffusion-model references.
Contemporaneous work broadens token corruption through Generalized
Interpolating Discrete Diffusion~\citep{vonrutte2025gidd}, moves diffusion to
latent segments for controllable generation~\citep{zhu2025segment}, or converts
pretrained autoregressive models to diffusion-style generation
\citep{cetin2025diffusion}. Most recently, DiffCoT~\citep{cao2026diffcot}
introduces a sliding window and causal noise schedule to jointly generate and
retrospectively revise reasoning steps; we include it as an advanced
same-backbone baseline. DiffCoT revises steps within an autoregressive chain,
whereas SRD constructs coherent states through typed, executable semantic
operators and learns their ordered local inverse transitions.
Diffusion language models learn text generation in token space, whereas SRD
learns semantic transitions in reasoning-state space; they share progressive
corruption and reverse recovery but differ in both state space and learning
objective.

\begin{figure*}[t]
\centering
\includegraphics[width=0.90\textwidth]{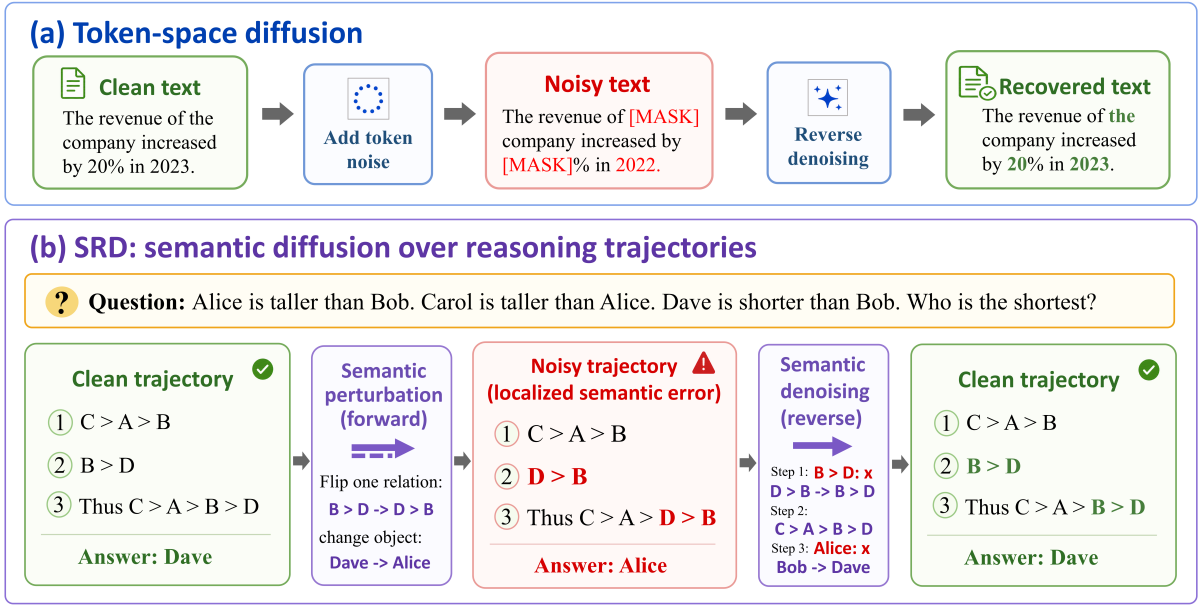}
\caption{Token-space and semantic denoising. \textbf{(a)} Token-space diffusion
corrupts surface tokens without modeling reasoning validity. \textbf{(b)} SRD
applies and inverts executable semantic operators over coherent reasoning
states to recover the trajectory and answer.}
\label{fig:example}
\end{figure*}


\section{Method}
\label{sec:method}

We introduce \textbf{Semantic Reasoning Denoising} (SRD), a discrete Markov
process over reasoning trajectories with transitions induced by semantic error
operators. Its forward process composes operators into corrupted states, and
its learned reverse process predicts and inverts the active operator
(\cref{fig:example}).

\subsection{Preliminaries}
\label{sec:method:setup}

Let a reasoning problem be a pair $(x, r_0)$, where $x$ is a question and $r_0$ a
\emph{correct} reasoning trajectory whose executed conclusion matches the
ground-truth answer, $\mathcal{E}(r_0)=y^\star$. We treat $r_0$ as the clean
reference state of the operatorized Markov denoising process. Let $\mathcal{Z}$
be a library of semantic error
operators. An operator $z=(\kappa,\ell,u,v)\in\mathcal{Z}$ records its error
type $\kappa$, step location $\ell$, original proposition $u$, and corrupted
proposition $v$. The typed library covers premise insertion/deletion, relation
reversal, intermediate-value substitution, step reordering, and conclusion
substitution.
We define the semantic-noise variable itself by this executable operator,
\begin{equation}
  \epsilon_t\coloneqq z_t
  =(\kappa_t,\ell_t,u_t,v_t).
  \label{eq:semantic_noise_operator}
\end{equation}
Here $\epsilon_t$ denotes semantic noise and $z_t$ its structured, executable
operator representation.
For a state $r_t$ produced by $t$ forward transitions, the ordered prefix
$z_{1:t}=(z_1,\ldots,z_t)$ represents its accumulated semantic noise, while
$z_t$ is the component inverted by the next adjacent reverse transition.
Its executable action $\mathcal{A}_z$ maps one coherent trajectory to another;
a paired repair action $\mathcal{R}_z$ removes that corruption. We do not require
every language edit to have a unique algebraic inverse, only that the recorded
pair satisfies $\mathcal{R}_z(\mathcal{A}_z(r))=r$ on constructed transitions.
For two trajectories $r$ and $r_0$, define their operator distance as
\begin{equation}
d_{\mathcal Z}(r,r_0)
=\min\!\left\{m:\exists z_{1:m},\
r=\mathcal A_{z_m}\circ\cdots\circ\mathcal A_{z_1}(r_0)\right\}.
\label{eq:operator_distance}
\end{equation}
This distance is the discrete counterpart of noise magnitude: it counts the
shortest executable corruption path rather than surface-token differences.

Continuous diffusion defines a Gaussian forward process
$q(r_t\mid r_0)=\mathcal{N}(r_t;\sqrt{\bar\alpha_t}\,r_0,(1-\bar\alpha_t)\mathbf{I})$
and learns a reverse chain $p_\theta(r_{t-1}\mid r_t)$
\citep{ho2020denoising,sohl2015deep}. In contrast, SRD uses compositions of
operators in $\mathcal{Z}$ to define explicit semantic-state transitions, without
relying on a Gaussian terminal distribution or a likelihood-based diffusion
objective.

\subsection{Forward Process}
\label{sec:method:forward}

Starting from $r_0$, the forward process samples and applies one operator (or a
small commuting bundle) at each step:
\begin{equation}
  z_t\sim q_\phi(\,\cdot\mid r_{t-1},x),\qquad
  r_t=\mathcal{A}_{z_t}(r_{t-1}),
  \label{eq:forward}
\end{equation}
Here, $q_\phi$ is the fixed empirical mixture induced by recovered real failures and
the synthetic operator sampler; it is not optimized jointly with the denoiser.
Together, these transitions define the Markov chain
\begin{equation}
\begin{aligned}
 q_\phi(r_{1:T},z_{1:T}\mid r_0,x)
 &=\prod_{t=1}^{T}q_\phi(z_t\mid r_{t-1},x)\\[-2pt]
 &\quad{}\cdot
 \mathbf{1}\!\left[r_t=\mathcal{A}_{z_t}(r_{t-1})\right].
\end{aligned}
 \label{eq:schedule}
\end{equation}
Each primitive operator is one semantic corruption step (a commuting bundle
counts as the number of its primitives). We therefore set
$\tau_t=t/T$ as the normalized process coordinate of the constructed forward
operator chain. It marks the relative position of a training state from the
clean start to the corrupted endpoint.

A clean trajectory is corrupted from three complementary sources represented
in the shared operator space:
\begin{enumerate}
  \item \textbf{Real failures (typically high $\tau$)}: the base model's own
    incorrect trajectories on training questions provide the closest samples
    of its natural error distribution. Aligning a failure with $r_0$ recovers
    its accumulated semantic noise as an ordered operator sequence.
  \item \textbf{Synthetic corruptions (controllable $\tau$)}: operators sampled
    from the library perturb an intermediate result, omit a premise, or alter a
    relation. Their number and composition directly control the process
    coordinate and provide exact inverse supervision.
  \item \textbf{Interpolated traces (intermediate $\tau$)}: applying only a
    prefix of a recovered or synthetic operator sequence constructs states
    between the clean trajectory and the complete erroneous trajectory.
\end{enumerate}
All three sources are represented as paired supervision between a corrupted
state $r_t$, its ordered noise operators $z_{1:t}$, and the correct trajectory
$r_0$. States at different process coordinates simulate the intermediate
stages encountered when progressively denoising a natural failure; each
transition pair uses $z_t$ as the next corruption component to invert.
These sources induce the training-state mixture
\begin{equation}
q_{\mathrm{mix}}
=\alpha q_{\mathrm{real}}
+\beta q_{\mathrm{syn}}
+\gamma q_{\mathrm{interp}},
\label{eq:training_mixture}
\end{equation}
where the coefficients are the empirical source proportions.
\paragraph{Operator extraction and executability checks.}
We split each failure and its clean trajectory into ordered reasoning steps,
align them monotonically, and map each mismatch to the smallest applicable
typed operator, including its location and corrupted/repaired payloads. We
retain only sequences whose execution reconstructs the paired states and whose
prefixes remain well formed; full construction details are given in the
supplementary material. Clean
($\tau\!=\!0$) identity pairs teach preservation. Since all states remain
coherent natural-language trajectories, a LoRA adapter only needs to learn the
structured mapping from corrupted states to local inverse edits.

\subsection{Reverse Process}
\label{sec:method:reverse}

\begin{algorithm}[t]
\caption{SRD inference for one problem $x$.}
\label{alg:infer}
\begin{algorithmic}[1]
\Require problem $x$, maximum denoising iterations $K$, denoiser $f_\theta$
\State $r \gets \textsc{BaseCoT}(x)$ \Comment{adapter disabled: initial state $r^{(0)}$}
\State $\tau\gets 1$
\For{$k = 1$ to $K$}
  \State $(\hat z,\hat r_{\mathrm{adj}}) \gets f_\theta(r,\,x,\,\tau)$
  \If{$\hat z=\varnothing$}
    \State \textbf{break}
  \EndIf
  \State $r'\gets\mathcal{R}_{\hat z}(r)$
  \If{$r'=r$} \State \textbf{break} \EndIf
  \State $r\gets r'$; \quad $\tau\gets\max(0,\tau-1/K)$
\EndFor
\State \Return $\mathcal{E}(r)$ \Comment{execute/extract the answer from the stable trace}
\end{algorithmic}
\end{algorithm}

The reverse process learns to invert the corruption. A single denoiser $f_\theta$
---an instruction-tuned LLM backbone with low-rank adapters
\citep{hu2022lora}---takes a corrupted trace $r_t$, the question $x$, and the
normalized process coordinate $\tau_t$ (serialized as a special \texttt{[t]} token,
mirroring instruction-guided editing \citep{brooks2023instructpix2pix}).
It jointly predicts the active operator $z_t$ and the supervised adjacent
lower-noise state $r_{t-1}^{\star}$ paired with $r_t$. The superscript
distinguishes this reconstruction label from the state actually produced at
inference. Adjacent-state prediction is an auxiliary transition-reconstruction
objective: it teaches recovery from semantic noise, whereas the predicted
inverse operator alone determines the next state actually entered. Formally,
the training output factorizes as
\begin{equation}
\begin{aligned}
&p_\theta(z_t,r_{t-1}^{\star}\mid r_t,x,\tau_t)\\
&=p_\theta(z_t\mid r_t,x,\tau_t)\,
  p_\theta(r_{t-1}^{\star}\mid r_t,x,\tau_t,z_t).
\end{aligned}
\label{eq:reverse_kernel}
\end{equation}
The operative state transition remains
$r_{t-1}=\mathcal R_{z_t}(r_t)$.

The model is trained with a joint supervised objective for structured operator
identification and adjacent-state reconstruction:
\begin{equation}
\begin{aligned}
  \mathcal{L}_{\mathrm{SRD}}
  &= \mathbb{E}_{x,r_t,z_t,r_{t-1}^{\star}} \Big[
      \underbrace{-\log p_\theta(z_t \mid r_t,x,\tau_t)}
      _{\textstyle \mathcal{L}_{\mathrm{op}}\ \text{(operator prediction)}}\\
  &\quad + \lambda
      \underbrace{\big(-\textstyle\sum_j w_j\log
      p_\theta(r_{t-1}^{\star(j)}\mid r_t,x,\tau_t,z_t)\big)}
      _{\textstyle \mathcal{L}_{\mathrm{adj}}\ \text{(adjacent-state prediction)}}
    \Big].
\end{aligned}
\label{eq:loss}
\end{equation}
The output sequence first specifies the current operator
$z_t=(\kappa_t,\ell_t,u_t,v_t)$ and then reconstructs
$r_{t-1}^{\star}$ conditioned on $r_t$ and $z_t$; $w_j$ upweights affected
positions. Removing either loss leaves only adjacent-state reconstruction or
operator diagnosis, respectively.

\begin{table*}[t]
\centering
\small
\renewcommand{\arraystretch}{1.06}
\begin{tabular}{llccccccc}
\toprule
\rowcolor{black!7}
Group & Method & GSM8K & MATH & HEval+ & MBPP & MMLU & PIQA & Avg. \\
\midrule
Base & CoT & 77.2 & 31.0 & 52.1 & 51.3 & 63.0 & 81.9 & 59.4 \\
\midrule
\multirow{3}{*}{Learn-to-solve}
 & SFT  & 79.3 & 35.6 & 55.5 & 52.4 & 65.5 & 82.0 & 61.7 \\
 & STaR & 80.6 & 39.8 & 54.7 & 52.9 & 66.3 & 84.2 & 63.1 \\
 & DiffCoT & 82.1 & 39.1 & 55.9 & 56.1 & 65.4 & 87.3 & 64.3 \\
\midrule
\multirow{3}{*}{Learn-to-correct}
 & Self-Refine & 80.4 & 34.5 & 52.8 & 51.6 & 62.7 & 83.5 & 60.9 \\
 & RISE & 82.3 & 38.5 & 53.2 & 52.0 & 68.9 & 83.4 & 63.1 \\
 & SCoRe & 80.6 & 36.7 & 53.5 & 55.4 & \textbf{70.1} & 86.1 & 63.7 \\
\midrule
\multirow{2}{*}{Reference}
 & Dream-7B & 77.2 & 39.6 & 57.9 & 56.2 & 69.5 & 75.8 & 62.7 \\
 & LLaDA-8B & 70.9 & 30.7 & 32.9 & 39.0 & 65.9 & 74.8 & 52.4 \\
\midrule
\rowcolor{black!4}
\textbf{Ours} & \textbf{SRD} & \textbf{84.8} & \textbf{43.1} & \textbf{57.9} & \textbf{58.4} & 69.6 & \textbf{91.0} & \textbf{67.5} \\
\bottomrule
\end{tabular}
\caption{Main in-domain results; diffusion-model rows are reference-only.}
\label{tab:exp1_main}
\end{table*}

\paragraph{Iterative denoising.}
Let $p_{\mathrm{base}}(r\mid x)$ denote the trajectory distribution induced by
the frozen backbone under the fixed prompting and decoding protocol. At
inference, we disable the adapter and draw the initial trajectory
$r_{\mathrm{init}}\sim p_{\mathrm{base}}(r\mid x)$, which we denote by
$r^{(0)}$. We then apply $f_\theta$ for at most $K$ iterations.
Here $K$ is a maximum denoising budget, not the unknown semantic-corruption
depth of the naturally generated initial trajectory. The schedule
$\tau^{(k)}=1-k/K$ traverses the same process coordinate in reverse.
At each step it predicts the current inverse operator together with an auxiliary
adjacent-state reconstruction:
\begin{equation}
\begin{aligned}
  (\hat z^{(k)},\hat r_{\mathrm{adj}}^{(k)})&=
  f_\theta(r^{(k)},x,\tau^{(k)}),\\
  r^{(k+1)}&=\mathcal R_{\hat z^{(k)}}(r^{(k)}),
  \qquad k=0,\dots,K-1 .
\end{aligned}
\label{eq:reverse_inference}
\end{equation}
We stop on an empty operator, an unchanged state, or at the step budget, then
read $\mathcal E(r)$.

For an exact operator posterior, reversing the terminal operator on a shortest
corruption path decreases $d_{\mathcal Z}$ by exactly one. This property
formalizes the unit-step geometry of the constructed reverse path and motivates
adjacent inverse-transition supervision; its proof is given in the
supplementary material.

At inference, every step after the first consumes a trajectory the denoiser
produced itself rather than one drawn from the constructed training
distribution, which could in principle expose the model to states it never saw
during training. We mitigate this by construction: the three noise sources jointly approximate
the severity and diversity of the backbone's own errors
(\cref{sec:exp:analysis}), so a self-produced intermediate trajectory likely
resembles the corruption space the denoiser was trained on. We find aggregate
accuracy remains stable in practice: after the first few steps, increasing the
denoising budget produces a performance plateau rather than a systematic
decline (\cref{sec:exp:efficiency}).


\section{Experiments}
\label{sec:exp}

We evaluate SRD on in-domain performance, cross-dataset transfer,
semantic-noise construction, sample and inference efficiency, and core
ablations.

\begin{table*}[!t]
\centering
\small
\renewcommand{\arraystretch}{1.06}
\begin{tabular}{llcccccccc}
\toprule
\rowcolor{black!7}
Backbone & Method & BBH & CSQA & WinoG. & BoolQ & ARC-C & SQuAD & QuAC & Avg. \\
\midrule
\multirow{2}{*}{Reference}
& Dream-7B & 59.4 & 80.0 & 63.0 & 45.0 & 68.0 & 66.1 & 15.9 & 56.8 \\
& LLaDA-8B & 47.4 & 71.2 & 73.5 & 41.2 & 47.5 & 59.7 & 13.2 & 50.5 \\
\midrule
\multirow{6}{*}{Llama-3}
& CoT (base) & 79.3 & 74.5 & 66.0 & 78.5 & 72.0 & 83.6 & 19.7 & 67.7 \\
& STaR & 83.4 & 79.6 & 72.0 & 87.0 & 78.4 & 85.7 & 21.4 & 72.5 \\
& DiffCoT & \textbf{83.7} & 79.5 & 81.1 & 84.0 & \textbf{84.1} & 88.1 & 24.0 & 74.9 \\
& RISE & 81.1 & 80.2 & 75.7 & 85.0 & 77.6 & 82.9 & 22.4 & 72.1 \\
& SCoRe & 80.5 & 81.4 & 80.2 & 85.4 & 80.3 & 84.5 & 20.8 & 73.3 \\
\rowcolor{black!4}
& \textbf{SRD (ours)} & 83.3 & \textbf{82.6} & \textbf{82.3} & \textbf{88.0} & 83.5 & \textbf{89.2} & \textbf{25.6} & \textbf{76.4} \\
\midrule
\multirow{6}{*}{Qwen3}
& CoT (base) & 77.8 & 82.5 & 67.5 & 89.5 & 90.5 & 85.5 & 17.3 & 72.9 \\
& STaR & 81.4 & 85.0 & 73.5 & 92.1 & 90.2 & 86.2 & 24.1 & 76.1 \\
& DiffCoT & 79.2 & 86.4 & 82.3 & 91.6 & 90.8 & 87.4 & 23.6 & 77.3 \\
& RISE & 79.1 & 85.5 & 80.8 & 90.7 & 88.9 & 86.9 & 25.7 & 76.8 \\
& SCoRe & 80.4 & 83.9 & 81.6 & 91.5 & 91.5 & 87.6 & 24.3 & 77.3 \\
\rowcolor{black!4}
& \textbf{SRD (ours)} & \textbf{85.2} & \textbf{89.8} & \textbf{84.7} & \textbf{92.4} & \textbf{93.7} & \textbf{89.0} & \textbf{26.4} & \textbf{80.2} \\
\bottomrule
\end{tabular}
\caption{Cross-dataset transfer results; reading tasks report F1.}
\label{tab:exp2_transfer}

\vspace{0.08in}
\begin{minipage}[t]{0.485\textwidth}
\centering
\includegraphics[width=\linewidth]{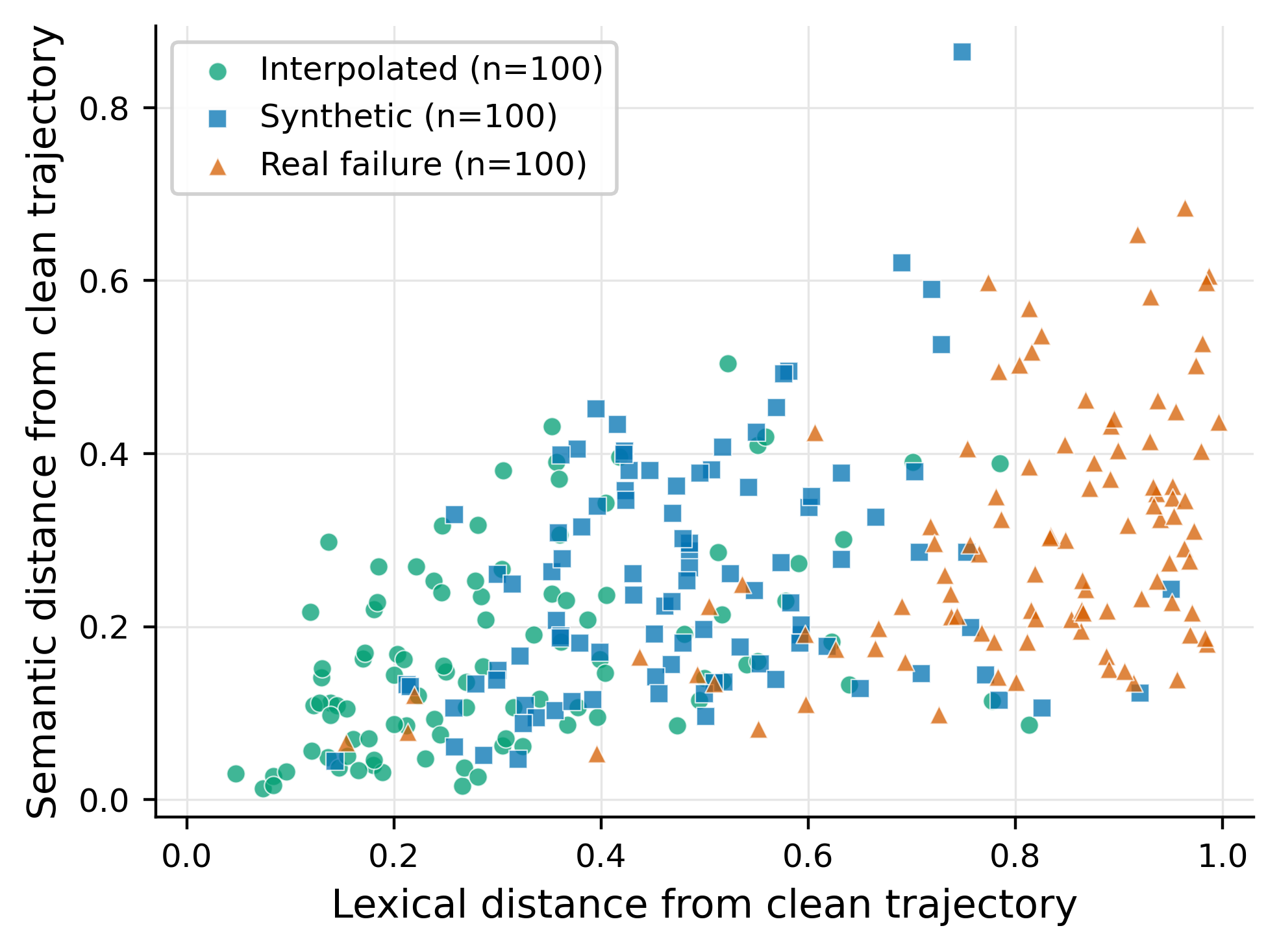}
\captionof{figure}{Semantic-noise sources in MATH (\(n=100\) each).}
\label{fig:noise}
\end{minipage}
\hfill
\begin{minipage}[t]{0.485\textwidth}
\centering
\includegraphics[width=\linewidth]{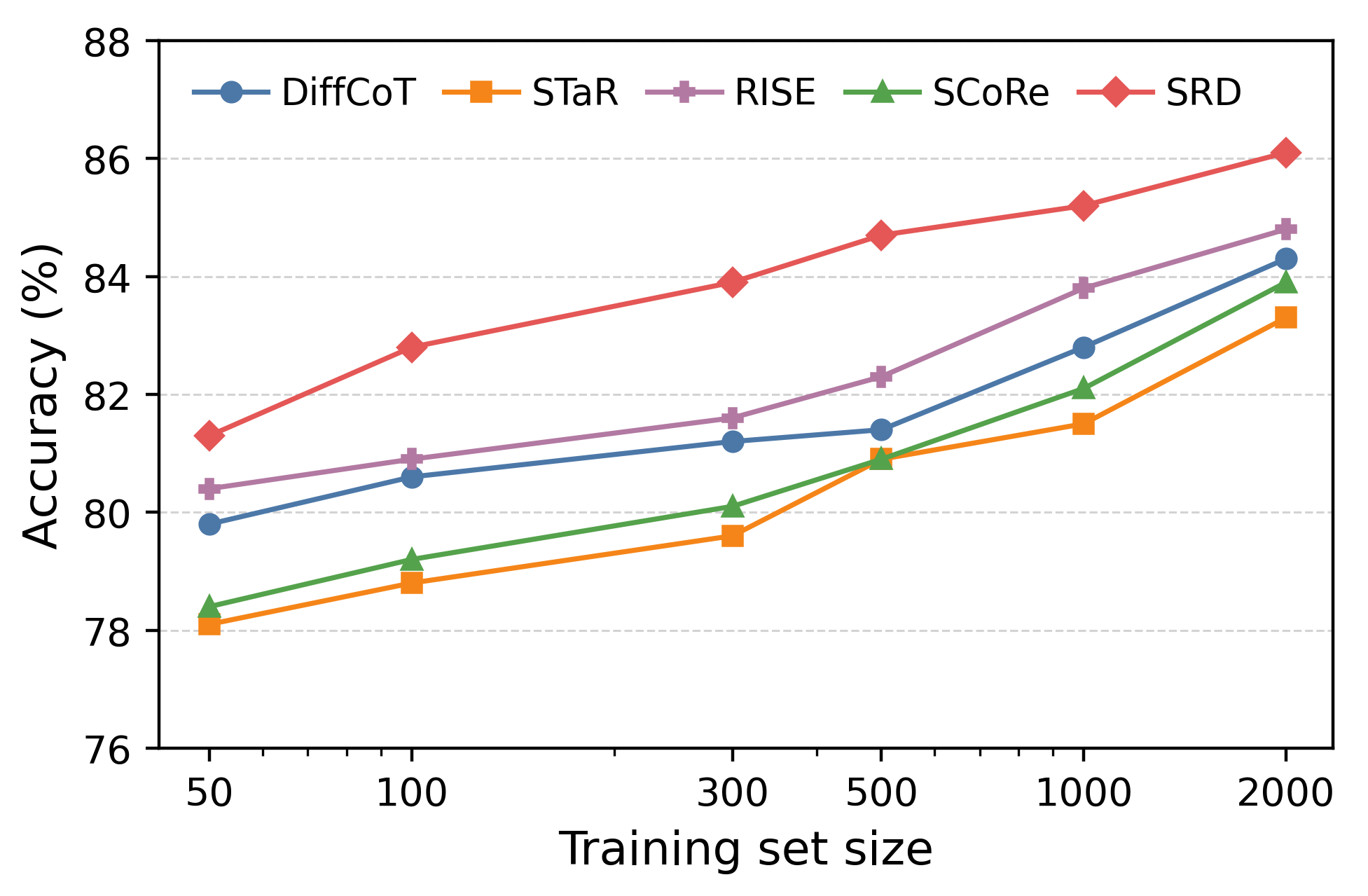}
\captionof{figure}{GSM8K accuracy versus training size (log-scale \(x\)-axis).}
\label{fig:sample_eff}
\end{minipage}
\end{table*}

\subsection{Main In-Domain Results}
\label{sec:exp:main}
\label{sec:exp:indomain}

\paragraph{Setup.}
We evaluate GSM8K, MATH, HumanEval+, MBPP, MMLU, and
PIQA~\citep{cobbe2021gsm8k,hendrycks2021math,liu2023evalplus,
austin2021mbpp,hendrycks2021mmlu,bisk2020piqa}.
All same-backbone methods use Llama-3-8B-Instruct and the corresponding training
split; because HumanEval+ has no supervised training split, we use MBPP for code
adaptation. Baselines follow their official papers and code. SRD transitions
$(x,r_t,z_t,r_{t-1}^{\star},\tau_t)$ come from aligned backbone failures,
executable synthetic operators, and operator-chain prefix states
(\cref{sec:method}). All trainable models use the checkpoint with the lowest
validation loss. For benchmarks with more than 1,000 test instances, every
method uses the same fixed 1,024-instance subset sampled with seed 42; otherwise,
we use the full test set. Test subsets are not used for model selection, and
main results are averaged over five independent random seeds.

\paragraph{Baselines.}
We compare with three families of methods. \emph{Learn-to-solve} baselines
(SFT, DiffCoT, STaR) improve reasoning through supervised trajectories,
self-generated rationales, or diffusion-styled reasoning optimization.
\emph{Learn-to-correct} baselines (Self-Refine, RISE,
SCoRe) revise an existing answer or reasoning trace. Dream-7B and LLaDA-8B are
included as diffusion-language-model references.

\paragraph{Metrics.}
We report accuracy for GSM8K, MATH, MMLU, and PIQA, and execution-based pass@1
for HumanEval+ and MBPP. Within each benchmark, all same-backbone methods use
the same prompt family, answer extractor, and metric implementation.

\Cref{tab:exp1_main} shows that SRD obtains the strongest overall performance
among same-backbone methods, with an average score of 67.5. The strongest
non-SRD baseline is DiffCoT, with an average of 64.3; SRD improves over it by 3.2
points on average, while also outperforming the strongest learn-to-correct
baseline SCoRe by 3.8 points. The advantage is most pronounced on tasks where
local reasoning errors directly determine the final answer: compared with STaR,
SRD is 4.2 points higher on GSM8K, 3.3 points higher on MATH, 5.5 points higher
on MBPP, and 6.8 points higher on PIQA. Across five independent runs per
method, SRD exceeds DiffCoT by 3.1 macro-average points; all six improvements
remain significant after Holm correction
(see the supplementary material). On MMLU, SCoRe is slightly
stronger than SRD, suggesting that factual multiple-choice questions may benefit
less from iterative trajectory denoising than math, code, and commonsense
reasoning. Compared with diffusion-language-model references, SRD matches
Dream-7B on HumanEval+ and is stronger on GSM8K, MBPP, and PIQA.

\subsection{Cross-Dataset Transfer}
\label{sec:exp:transfer}

\paragraph{Setup.}
The second experiment evaluates whether a learned corrector transfers beyond
the dataset used to build its supervision. Unlike the in-domain setting, we
construct source data by transfer group: ProofWriter for BBH; StrategyQA and
QASC for CSQA, WinoGrande, BoolQ, and ARC-C; and HotpotQA for SQuAD and
QuAC~\citep{tafjord2021proofwriter,suzgun2023bbh,geva2021strategyqa,
khot2020qasc,talmor2019commonsenseqa,sakaguchi2020winogrande,clark2019boolq,
clark2018arc,yang2018hotpotqa,rajpurkar2016squad,choi2018quac}.
SRD uses the same operator-based semantic-noise construction and joint
operator/adjacent-state objective in each group. We evaluate two
strong learn-to-solve baselines (DiffCoT, STaR) and two strong learn-to-correct
baselines (RISE, SCoRe) on both Llama-3-8B-Instruct and
Qwen3-8B~\citep{yang2025qwen3}. For Qwen3, we
disable thinking mode so that all methods use the same answer extractor.

\Cref{tab:exp2_transfer} evaluates whether the learned correction behavior
transfers beyond the dataset on which it was trained. On Llama-3, SRD improves
the base average from 67.7 to 76.4 (+8.7 points), with particularly large gains
on WinoGrande and ARC-C. SRD is best on CSQA, WinoGrande, BoolQ, SQuAD, and QuAC,
and obtains the highest Llama average. On
Qwen3-8B, SRD leads every reported target and improves the base average from
72.9 to 80.2 (+7.3 points). This two-backbone pattern supports cross-dataset and
cross-backbone transfer of the learned correction behavior.

\subsection{Semantic-Noise Analysis and Sample Efficiency}
\label{sec:exp:analysis}

We analyze SRD's semantic corruption space and sample efficiency using the
construction in \cref{sec:exp:main}.

\Cref{fig:noise} shows how the three noise sources populate the corruption
space. Lexical distance is one minus \texttt{SequenceMatcher} similarity;
semantic distance is one minus cosine similarity between MiniLM embeddings.
Interpolated trajectories are the mildest and synthetic corruptions have
moderate, controllable severity. The model's real failures are the most diverse,
with the heaviest tail toward large lexical and semantic distances. Their
overlapping ranges create a gradual severity continuum rather than disconnected
clusters: interpolation and synthetic corruption stabilize mild reverse
transitions, while real failures better approximate the natural errors
encountered during inference.

\Cref{fig:sample_eff} shows that SRD is consistently more sample-efficient than
the correction and self-training baselines. Its advantage is already visible
with 50 training examples and persists as the training set grows to 2,000,
indicating that learning structured semantic reverse transitions provides a
useful inductive bias rather than merely benefiting from additional data.

\subsection{Inference Efficiency}
\label{sec:exp:efficiency}

This analysis measures how the cost of SRD scales with the number of denoising
steps, using the same corrector and denoising procedure as \cref{sec:exp:main}
and varying only the number of steps. Measurements use BF16 precision and
greedy decoding on a single NVIDIA RTX 4090 D GPU.

\begin{table}[t]
\centering
\small
\begin{tabular}{ccccc}
\toprule
\rowcolor{black!7}
$K$ & Calls & Gen.\ tok. & Latency (s) & Tok./fwd. \\
\midrule
1  & 2  & 304  & 5.7  & 1.7 \\
2  & 3  & 445  & 6.4  & 2.5 \\
4  & 5  & 735  & 7.1  & 4.2 \\
8  & 9  & 1366 & 9.5  & 5.4 \\
16 & 17 & 2378 & 11.2 & 8.0 \\
\bottomrule
\end{tabular}
\caption{MATH inference cost by denoising budget.}
\label{tab:efficiency}
\end{table}

\Cref{tab:efficiency} reports the cost of SRD across denoising budgets. As $K$
increases from $1$ to $16$, the generated tokens grow by $7.8\times$ but the total
latency grows only $2.0\times$. Self-speculative decoding verifies matching draft
tokens in parallel and reuses shared-prefix KV caches, amortizing repeated
decoding of unchanged content and keeping iterative denoising practical.

\begin{figure}[t]
\centering
\includegraphics[width=\columnwidth]{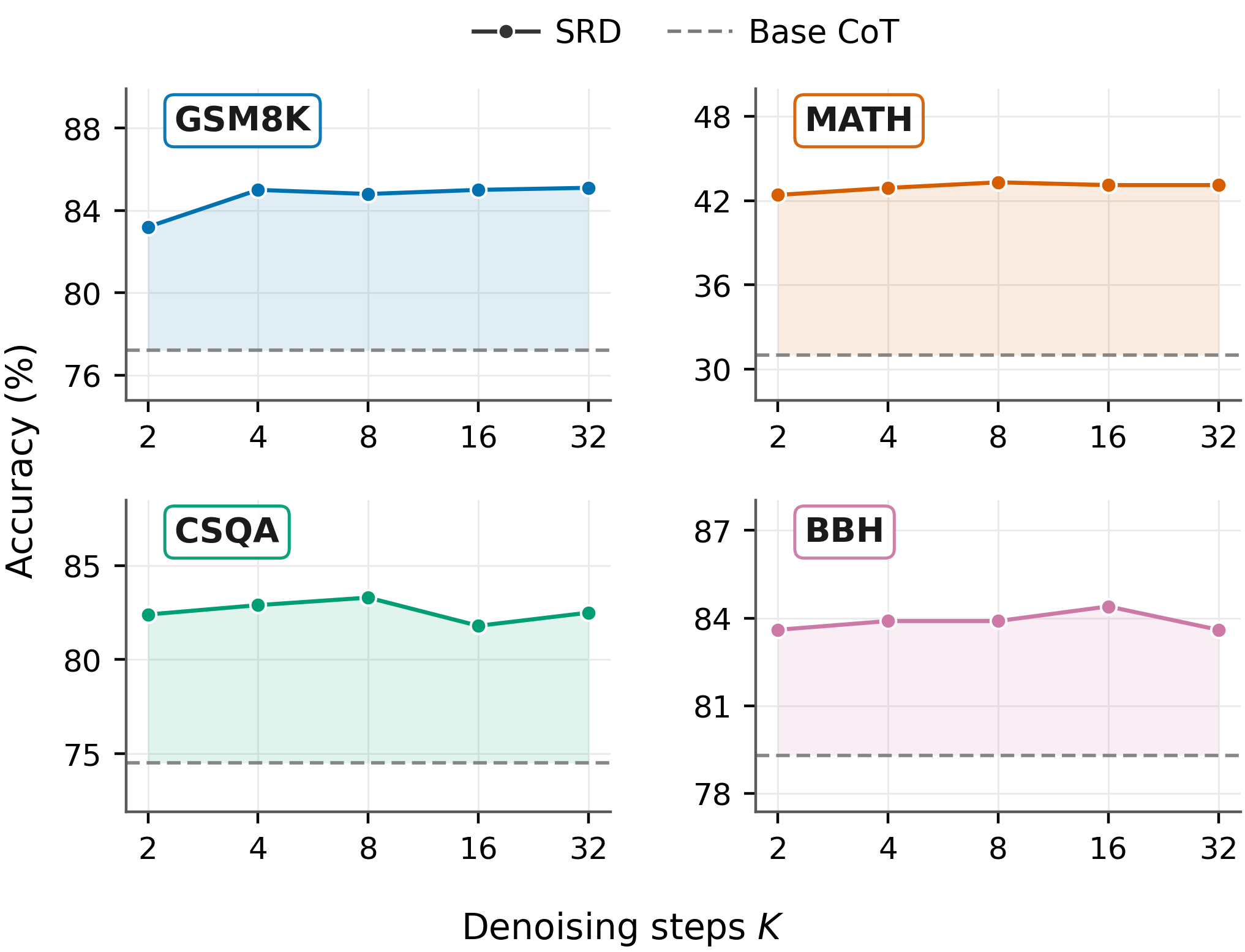}
\caption{Accuracy versus denoising budget $K$ on GSM8K, MATH, CSQA, and BBH;
dashed lines show base CoT.}
\label{fig:ksweep}
\end{figure}

\Cref{fig:ksweep} shows that SRD improves over base at every tested $K$, with
most gains realized in the first few steps. Larger budgets add cost
(\cref{tab:efficiency}) with little accuracy benefit, supporting a small default
$K$.

\subsection{Core Ablations}
\label{sec:exp:ablation}

We ablate SRD's noise sources, training objective, and supervision target,
changing only the component named in each row.

\begin{table}[t]
\centering
\small
\begin{tabular}{@{}lcccc@{}}
\toprule
\rowcolor{black!7}
Variant & GSM8K & MATH & CSQA & BBH \\
\midrule
\rowcolor{black!4}
\textbf{Full SRD} & \textbf{84.8} & 43.1 & \textbf{82.6} & \textbf{83.3} \\
\midrule
w/o real failures & 83.1 & 43.6 & 81.7 & 82.7 \\
w/o synthetic     & 80.4 & 41.4 & 81.2 & 82.1 \\
w/o interpolation & 82.7 & \textbf{44.5} & 81.6 & 82.1 \\
\midrule
operator-only       & 79.1 & 40.8 & 80.9 & 81.8 \\
adjacent-state-only & 80.6 & 40.3 & 81.2 & 82.4 \\
\midrule
answer-only supervision & 81.4 & 41.5 & 80.5 & 81.6 \\
full-CoT supervision    & 77.5 & 40.8 & 81.9 & 83.1 \\
\bottomrule
\end{tabular}
\caption{Core ablations (best per column in \textbf{bold}).}
\label{tab:ablation}
\end{table}

\Cref{tab:ablation} shows that full SRD is the most robust configuration.
Removing a noise source generally hurts, with synthetic noise contributing most
overall. Operator-only and adjacent-state-only objectives weaken all four tasks,
confirming their complementarity; answer-only and full-CoT supervision also
underperform semantic trajectory denoising.

\paragraph{Compositional operator diagnosis.}
We test one-, two-, and three-operator corruptions across all six in-domain
benchmarks; full metrics and sample counts appear in the supplementary material.
SRD retains strong parsing and operator recognition as depth increases. With
three operators, execution reaches 81.4\% on GSM8K and 85.7\% on PIQA, while
answer recovery remains 77.0\% and 84.6\%, respectively. Performance declines
more on MATH and code, identifying localization as the main bottleneck for
longer compositions.


\section{Conclusion}
\label{sec:conclusion}

We presented SRD, an operatorized Markov denoising method that learns executable
inverse transitions over reasoning trajectories. SRD improves the strongest
same-backbone baseline by 3.2 points and transfers across Llama-3 and Qwen3.
Ablations and compositional diagnostics support joint operator and adjacent-state
supervision, while identifying localization as the bottleneck for longer chains.

\paragraph{Limitations and future work.}
Evaluation is limited to 8B backbones and fixed operators, motivating larger
models and adaptive operators for long or ambiguous error chains.

\bibliographystyle{aaai2027}

\end{document}